\documentclass[letterpaper]{article} 
\usepackage[]{aaai24}  
\usepackage{times}  
\usepackage{helvet}  
\usepackage{courier}  
\usepackage[hyphens]{url}  
\usepackage{graphicx} 
\usepackage{natbib}  
\usepackage{caption} 
\usepackage{algorithm}
\usepackage{algorithmic}
\usepackage{booktabs}

\usepackage{amsmath, bm}
\usepackage{amssymb}

\usepackage{newfloat}
\usepackage{listings}
\DeclareCaptionStyle{ruled}{labelfont=normalfont,labelsep=colon,strut=off} 
\floatstyle{ruled}
\newfloat{listing}{tb}{lst}{}
\floatname{listing}{Listing}
\title{Challenges of Data-Driven Simulation of Diverse and Consistent \linebreak Human Driving Behaviors}
\author{
    Kalle Kujanpää\textsuperscript{\rm $\ast$}\textsuperscript{\rm $\ddagger$}
    Daulet Baimukashev\textsuperscript{\rm $\dagger$}
    Shibei Zhu\textsuperscript{\rm $\ast$}\textsuperscript{\rm $\ddagger$}
    Shoaib Azam\textsuperscript{\rm $\dagger$}\textsuperscript{\rm $\ddagger$}
    Farzeen Munir\textsuperscript{\rm $\dagger$}\textsuperscript{\rm $\ddagger$} \\
    Gokhan Alcan\textsuperscript{\rm $\dagger$}
    Ville Kyrki\textsuperscript{\rm $\dagger$}\textsuperscript{\rm $\ddagger$}
}
\affiliations{
(kalle.kujanpaa, daulet.baimukashev, shibei.zhu, shoaib.azam, farzeen.munir, gokhan.alcan, ville.kyrki)@aalto.fi \\
    \textsuperscript{\rm $\ast$}Department of Computer Science
    \textsuperscript{\rm $\dagger$}Department of Electrical Engineering and Automation\\
    \textsuperscript{\rm $\ddagger$}Finnish Center for Artificial Intelligence FCAI\\
    Aalto University

}

\usepackage{bibentry}
\usepackage{amsthm}
\usepackage{amsmath}

\begin{document}

\maketitle

\begin{abstract}
Building simulation environments for developing and testing autonomous vehicles necessitates that the simulators accurately model the statistical realism of the real-world environment, including the interaction with other vehicles driven by human drivers. To address this requirement, an accurate human behavior model is essential to incorporate the diversity and consistency of human driving behavior. We propose a mathematical framework for designing a data-driven simulation model that simulates human driving behavior more realistically than the currently used physics-based simulation models.
Experiments conducted using the NGSIM dataset validate our hypothesis regarding the necessity of considering the complexity, diversity, and consistency of human driving behavior when aiming to develop realistic simulators. 
\end{abstract}

\section{Introduction}
Modeling human behavior is pivotal in building intelligent learning systems, offering profound insights into how humans interact, behave, and respond within complex environments. This understanding is particularly crucial in the context of autonomous vehicle (AV) simulation \cite{feng2021intelligent}. Modeling and simulating human behaviors for building a data-driven simulation environment for autonomous vehicles is challenging due to human behaviors' high variability and diversity \cite{fuchs2022modeling, chen2023data}. As the simulation provides a cost-effective way to evaluate the performance of autonomous vehicles, achieving statistical realism in such a simulation environment remains a persistent challenge in the field, necessitating sophisticated modeling techniques to incorporate human behaviors.
\par 
A key component of creating a high-fidelity driving environment is precisely modeling human behaviors. There have been notable efforts in the literature to develop autonomous driving simulators, propelled by rapid advancements in artificial intelligence, computer graphics and vision, and high-performance computing devices. Simulators like CARLA \cite{dosovitskiy2017carla}, Airsim \cite{shah2018airsim}, DriveSim \footnote{\url{https://developer.nvidia.com/drive/simulation}}, Wayve Infinity Simulator \footnote{\url{https://wayve.ai/thinking/introducing-wayve-infinity-simulator/}}, and SimulationCity \footnote{\url{https://waymo.com/blog/2021/06/SimulationCity.html}} provide models for accurate vehicle dynamics and photo-realistic scenarios. However, these simulators mainly focus on vehicle dynamics and often overlook the driving environment, particularly the behavior of non-player characters, such as background road users. The behavior of these background vehicles is typically either replayed from logs or simulated using heuristic rules, creating a disparity between the simulation and the real-world driving environment. Furthermore, the microscopic simulation environments mimic the interaction between agents using physics-driving models, and hand-crafted rules such as car-following \cite{treiber2000congested}, gap-acceptance \cite{mahmassani1981using}, and lane-changing models \cite{kesting2007general, erdmann2015sumo} often lack the diversity to model the human behavior for simulated agents. Therefore, these physics and theory-based driving models can hardly be generalized and scaled to capture the underlying diversity of human behavior for driving.
\par
To this end, in this work, we focus our studies on building a high-fidelity simulation environment statistically representative of the real-world driving environment by modeling human driving behaviors. We aim to develop a data-driven simulation model that captures the subtle complexities and the wide range of variability inherent in human driving behavior in various scenarios and conditions. In this work, we raise the following research questions open to discussion:
\begin{enumerate}
\item \textbf{How can we accurately model the inherently unpredictable and stochastic nature of real human driving behavior?}
\item \textbf{Considering the diverse nature of human driving behavior, how can this diversity and variability, especially in different human driving styles, be effectively incorporated into the simulation environment?}
\item \textbf{How to incorporate the temporal consistency of human driving behavior in simulations?}
\end{enumerate}

\section{Related Work}
Human driving behavior is extensively studied, encompassing a broad range of problem definitions, assumptions in modeling, and diverse methodologies related to modeling human driving behavior \cite{brown2020modeling}. Driving is a widespread yet intricate dynamic activity that requires the integration of physiological, mental, and physical aspects in managing the vehicle to achieve a specific objective. It involves a repetitive interplay between the driver, the vehicle, and the traffic environment. The overall task of driving encompasses a wide range of physiological constructs, behaviors, and action patterns. Modeling the complex process of how humans execute diverse road maneuvers is challenging. The variability and diversity in human drivers add to the complexity of developing Driver Behavior Models (DBMs) \cite{negash2023driver}. 
\par 
Most classical DBMs such as the Intelligent Driver Model (IDM) \cite{treiber2000congested}, the MOBIL model \cite{kesting2007general}, and models by Gipps \cite{gipps1981behavioural,gipps1986model} are deterministic. These models often fail to encapsulate the stochastic nature inherent in high-fidelity human driving behavior. In these models, the parameters  are static during simulations, and the noise  is typically disregarded. Despite the calibration of parameter values with real-world data \cite{sangster2013application,li2016global,punzo2012can,ma2020sequence}, these deterministic models struggle to accurately reflect the variability and diversity in human driving behavior.
\par
Recently, researchers have shifted their focus towards employing machine learning-based methods \cite{xie2019data,huang2018car,wang2017capturing}, particularly neural networks, for fitting the DBM. Large naturalistic driving datasets are utilized to estimate the parameters that reflect observed trajectories of human drivers. However, they still lack accurate stochasticity. 
Introducing the noise into the model can integrate a degree of stochasticity. Typically, this noise is modeled as Gaussian noise \cite{kuefler2017imitating}. However, Gaussian noise may not adequately capture the variability and diversity inherent in human driving behavior. This is because the interactions in various driving conditions are highly complex and do not always conform to the simple Gaussian distribution \cite{laval2014parsimonious,treiber2017intelligent}. In \cite{yang2010development}, extreme-value and log-normal distributions are applied. However, the driving behavior exhibited by humans in different scenarios can vary significantly, thus necessitating a more sophisticated approach to modeling noise that accurately reflects this complexity.
\par
In addition to incorporating external noise, model parameters can also be used to introduce stochasticity. \citet{yang2010development} accounted for human errors in car-following by modeling time delay and distraction intervals using exponential and lognormal distributions, respectively. Game theory was applied in \citet{talebpour2015modeling} for stochastic lane-changing behavior modeling. \citet{hamdar2015behavioral} developed a utility-based stochastic car-following model with acceleration probabilities following a continuous logit model. However, these methods often overlook the distributional accuracy of driving behaviors post-introduction of stochasticity. Exceptions are found in \citet{wang2009markov} and \citet{chen2010markov}, where stochastic car-following models were proposed to accurately capture the distribution of time headway accurately, aligning simulation environments with real-world data. Nevertheless, these studies mainly focus on single-lane road scenarios and do not address error accumulation, which could significantly distort long-term simulations. 
\par
Despite various related studies, most existing methodologies fall short in accurately capturing the distribution of stochastic human driving behaviors. Consequently, the driving environments generated by these behavior models are not sufficiently precise in replicating the statistical characteristics of real-world simulation environments, an aspect crucial for autonomous vehicle (AV) testing.

\section{Problem Formulation}

Given a set of $n$ demonstrations collected from the real world as $\mathcal{D} = \{ \tau_0, \tau_1, \dots, \tau_n \}$, where each trajectory $\tau_i$ is given as a temporal sequence of state and action pairs $\tau_i = \{ (s_0, a_0), \dots, (s_T, a_T) \}$, the driver behavior modeling problem can be formulated as follows:
\begin{equation}
	\label{prob-formulatiion}
	\begin{aligned}
		\text{min} \quad & \mathop{\mathbb{E}}_{\tau_i \sim \mathcal{D}} \Bigg(J(\tau_i, \hat{\tau}_i)\Bigg) \\
		\text{subject to} \quad & \hat{\tau}_i = \{ (s_0, \hat{a}_0), \dots, (s_T, \hat{a}_T) \}, \\ 
		& \hat{a}_t \sim \pi_\theta(s_t,\psi_i), \\
        & \psi_i \sim p(\psi). \\
	\end{aligned}
\end{equation}
Here, $J(\tau_i, \hat{\tau}_i)$ is an objective function measuring the deviation between an actual trajectory $\tau_i$ and a predicted trajectory $\hat{\tau}_i$. $\psi$ represents the parameters of the driver behavior model, drawn from some distribution $p(\psi)$, and $\theta$ stands for the general parameters of the policy $\pi$.
\par
Eq.\ref{prob-formulatiion} represents the general form of modeling the human driving behavior for the simulated environment. Following this, two possible types of solutions arise that address the underlying question of \textbf{how to model the human driving behavior}: theory-based and data-driven models.
\par 
In theory-based methods, the policy parameters $\theta$ are generally not optimized. Rather, they are implicitly determined by the chosen driving model, such as IDM or its enhancements. Instead, to simulate human driving, the objective function $J(\tau_i, \hat{\tau}_i)$ can be minimized by learning the distribution $p(\psi)$.

Alternatively, when utilizing generic data-driven policies that are not structured as theory-based methods, such neural networks, there is an enhancement in generalization capabilities. In such instances, actions are often drawn from a policy directly parameterized by $\theta$, as follows:
\begin{equation}
    \hat{a}_t \sim \pi_{\theta}(s(t), \psi)
\end{equation}
For each time step $t$, the predicted action $\hat{a}_t(\theta)$ is determined by the policy function $\pi_\theta$, parameterized by $\theta$, based on the current state $s(t)$ and a set of parameters $\psi$. These parameters $\psi$ play a pivotal role in the model, as they can be either predefined priors or latent variables. This distinction allows the model to either leverage existing knowledge about driving behaviors (when $\psi$ is a prior) or to infer and optimize these parameters based on the observed data (when $\psi$ is treated as a latent variable). We argue that treating the parameters $\psi$ correctly is essential for accurately modeling the diversity of actual human behavior, and they should be driver-specific.
In the following case study, we compare a theory-based model, IDM, to a generic data-driven model in terms of accurately replicating human driving behavior.

\section{Case Study}

\begin{figure*}[t]
\centering
\includegraphics[width=\textwidth]{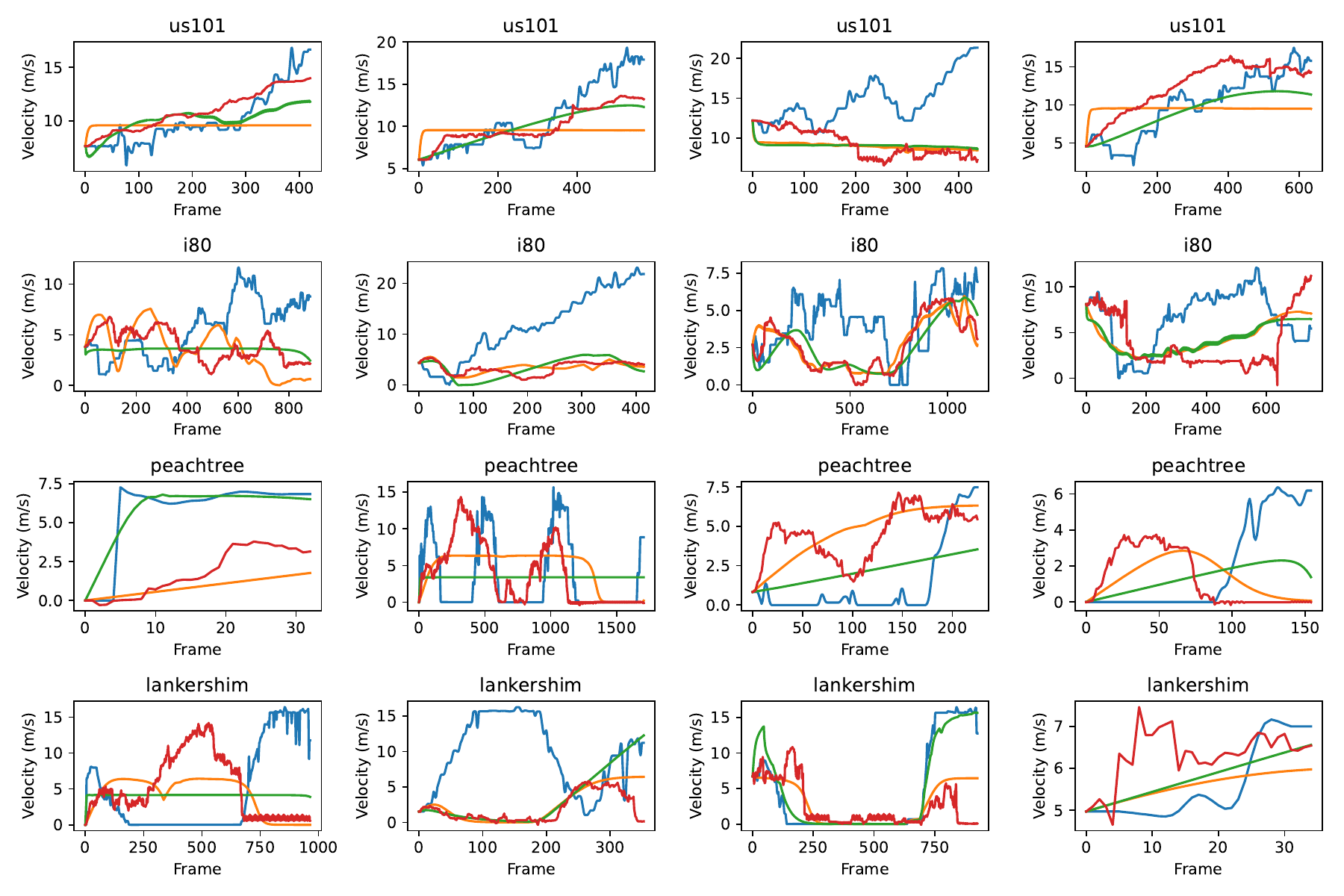}
\includegraphics[width=0.9\textwidth,trim={2.3cm 6mm 2.3cm 6mm},clip]{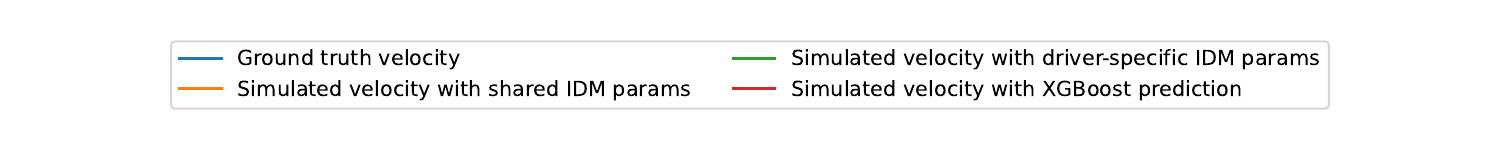}
\caption{We compare the velocity predictions of the two types of IDMs and a learned XGBoost predictor to the ground truth data in the simulated NGSIM environments. We find that replicating the human driving behavior is very difficult, even in this simplistic car-following setting.}
\label{fig:ngsim_idm_velocity}
\end{figure*}

\begin{figure*}b]
\centering
\includegraphics[width=\textwidth]{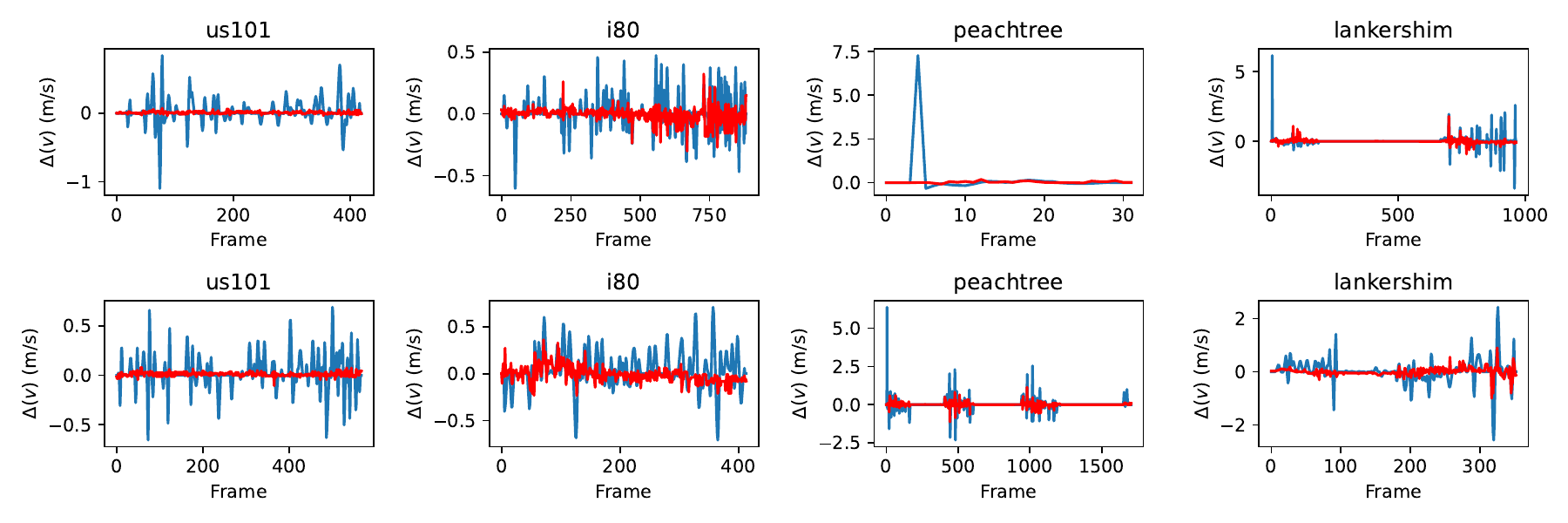}
\includegraphics[width=0.9\textwidth,trim={2.3cm 6mm 2.3cm 6mm},clip]{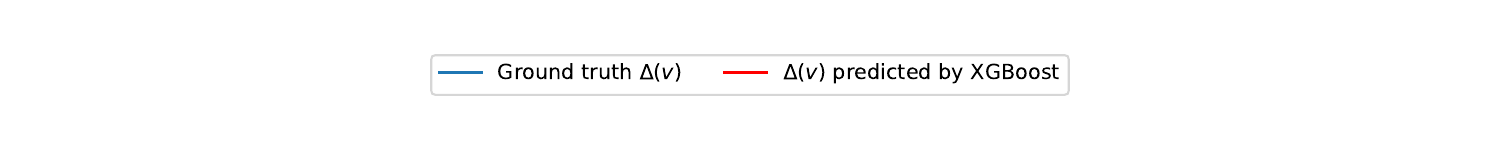}
\caption{We compare the change in velocity predicted by XGBoost to the ground truth change in velocity. The plot indicates that human behavior may be inherently unpredictable, and capturing the behavior seems to be highly challenging.}
\label{fig:xgb_onestep_preds}
\end{figure*}

\begin{table}
\centering
\begin{tabular}{l|l|c|c}
 Dataset & Parameters & MSE & SD \\
 \midrule
 US 101 & Driver-specific & 215.63 $\pm$ 3.08 & 240.55 \\
 & Shared & 289.44 $\pm$ 3.16 & 246.58 \\
 \midrule
 I-80 & Driver-specific & 220.16 $\pm$ 5.15 & 387.92 \\
 & Shared & 301.19 $\pm$ 5.82 & 438.52 \\
 \midrule
 Peachtree & Driver-specific & 58.50 $\pm$ 2.48 & 119.64 \\
 & Shared & 107.49 $\pm$ 3.27 & 157.66 \\
  \midrule
 Lankershim & Driver-specific & 264.19 $\pm$ 6.08 & 300.14 \\
 & Shared & 416.55 $\pm$ 6.14 & 303.25 \\
\end{tabular}
\caption{The mean squared error plus-minus standard error of the MSE for IDM models fitted to the NGSIM data. The results show that fitting driver-specific models outperforms using shared models, but the performance of both variants is highly suboptimal.}
\label{tab:fit_mse}
\end{table}

\begin{figure*}[h]
\centering
\includegraphics[width=\textwidth]{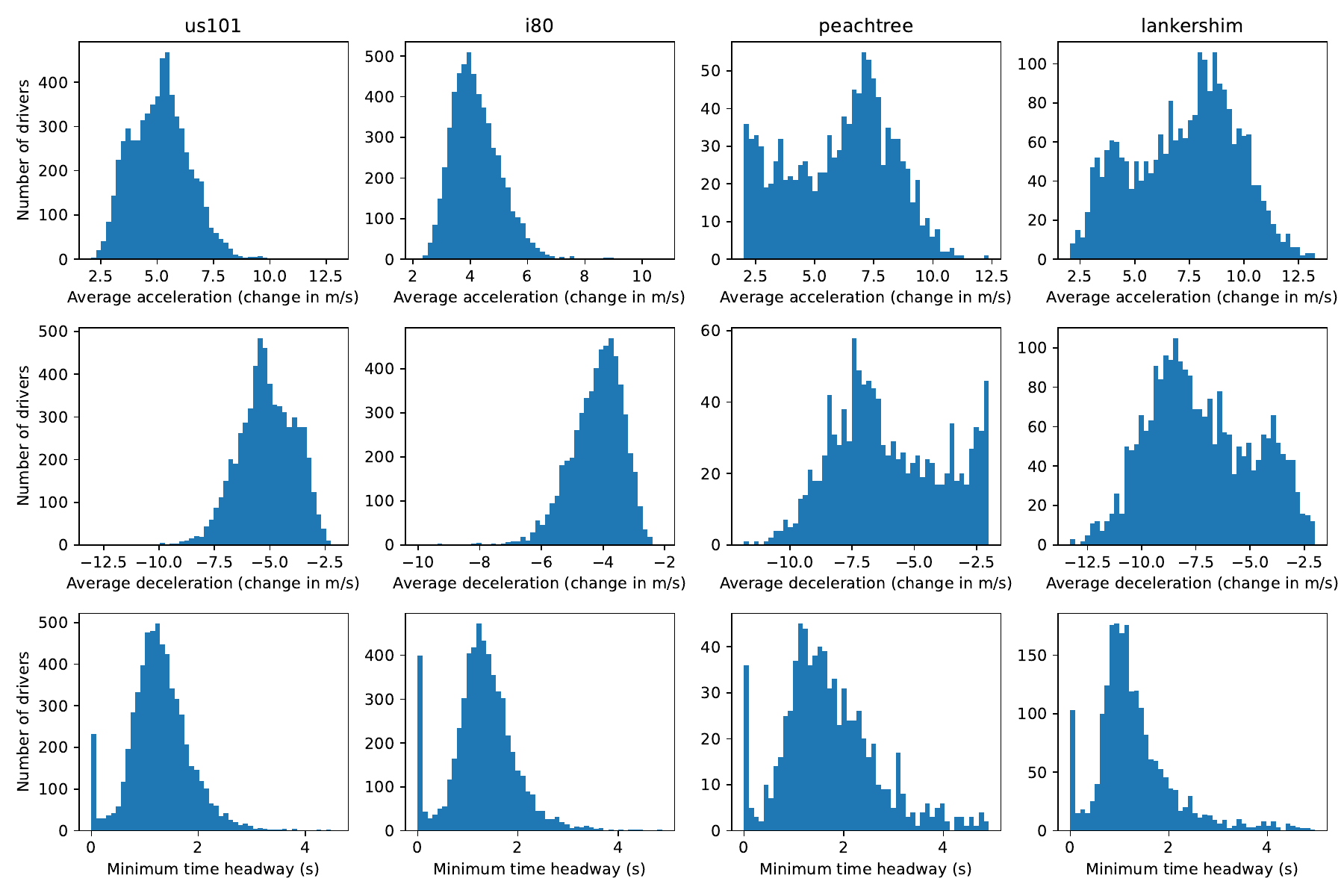}
\caption{We plot the average acceleration and deceleration and the preferred minimum time headway, that is, the safety margin between two vehicles for each driver in each of the four NGSIM datasets. The plots show that human driving behavior is highly diverse. It can be fat-tailed and multi-modal and does not simply follow a Gaussian distribution.}
\label{fig:ngsim_diversity}
\end{figure*}

\begin{figure*}[h]
\centering
\includegraphics[width=\textwidth]{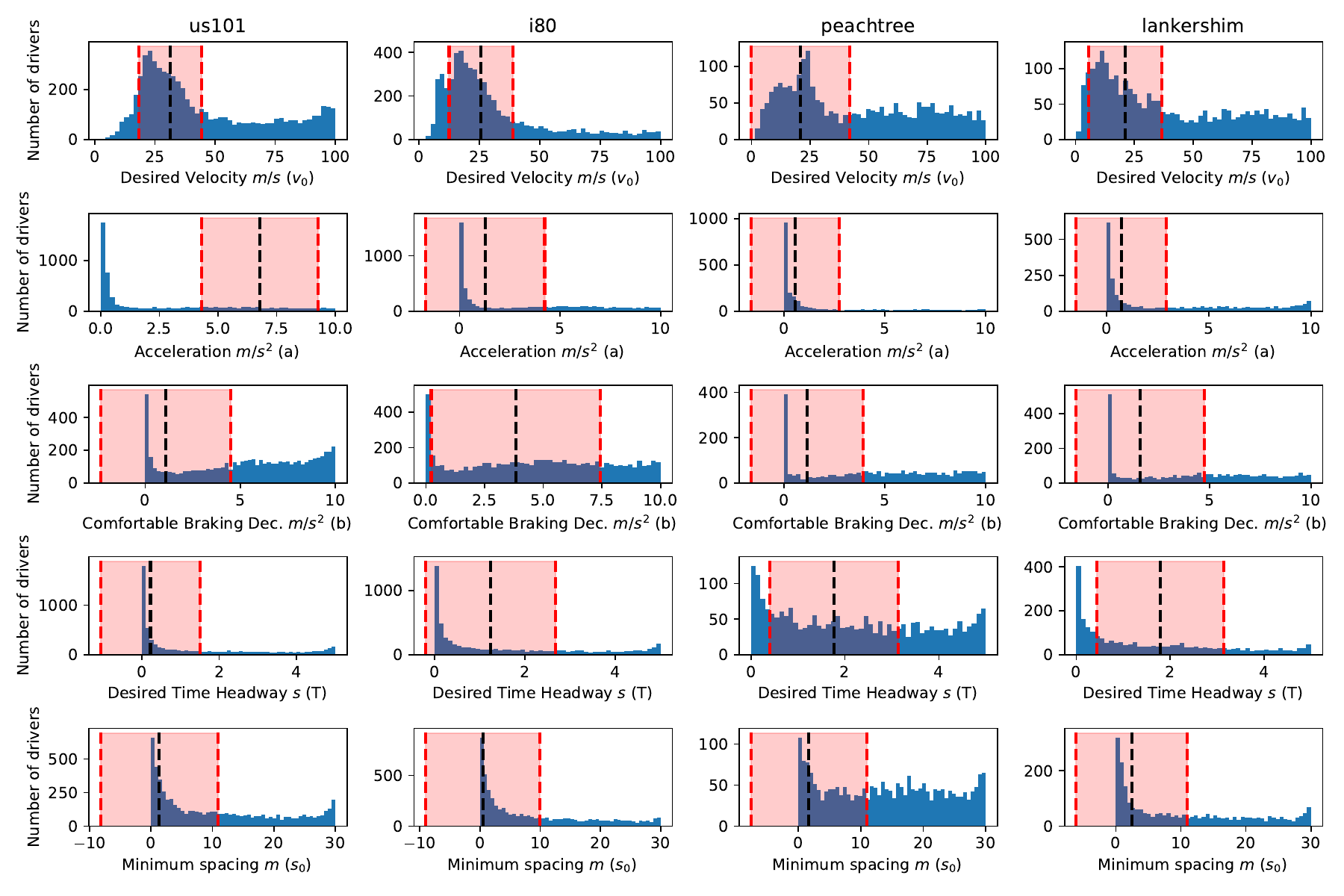}
\caption{We plot the distribution of the IDM parameters for each of the four NGSIM datasets when each driver is represented by a separate set of parameters. The black vertical line can be interpreted as the human mean, whereas the red-shaded area depicts the expected noise in the IDM fitting process.}
\label{fig:ngsim_idm_diversity}
\end{figure*}

\begin{figure*}[t]
\centering
\includegraphics[width=\textwidth]{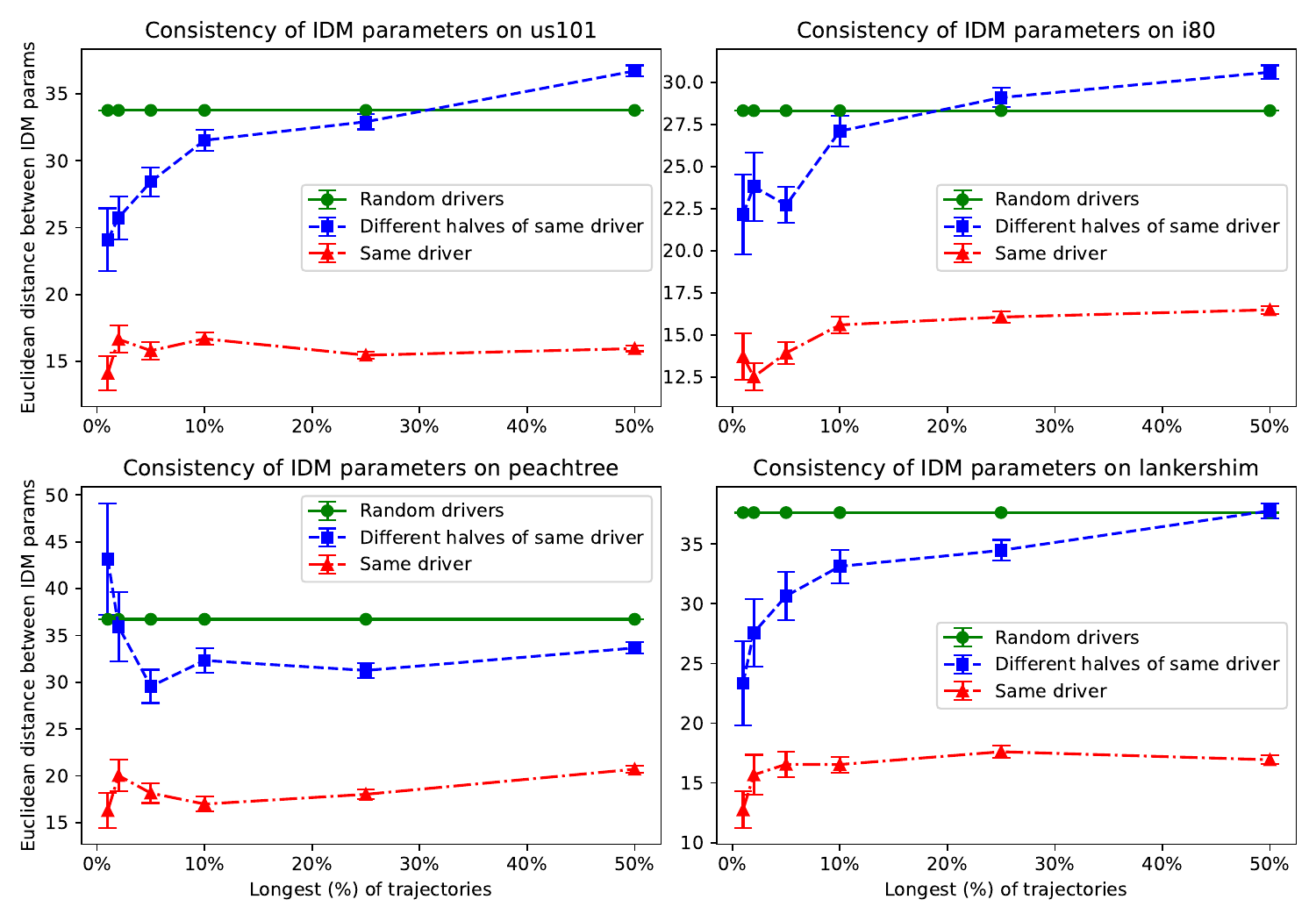}
\caption{We analyze the consistency of human drivers on the four datasets in the NGSIM data by fitting IDM to the longest trajectories in the dataset. Our results show that in all of the NGSIM datasets, the human driving style shows statistically significant consistency when we focus on the human drivers of which most data has been collected. The uncertainty plotted with error bars is one standard error of the mean.}
\label{fig:ngsim_consistency}
\end{figure*}

In this section, we present a series of experiments designed to highlight the importance and relevancy of the research questions outlined in the introduction. We discuss the modeling of human driving behavior for its realistic simulation and argue that human behavior is complex and diverse but also consistent. Utilizing the Next Generation Simulation (NGSIM) dataset as our foundation, we focus on the application of the Intelligent Driver Model (IDM) to predict the diverse driving behaviors observed in the dataset. Additionally, we scrutinize the limitation of a more robust model, XGBoost, in capturing realistic driving behavior. The results not only highlight the inherent diversity and stochastic nature of human driving styles but also underscore the importance of employing sophisticated methods capable of robustly addressing both the complexities and consistencies inherent in human behavior.

\subsection{Background}

\subsubsection{Intelligent Driver Model}
While modeling complex human driving is challenging, researchers have proposed various approaches to tackle, for instance, car-following and lane-changing behaviors. One prominent model in this domain is the Intelligent Driver Model (IDM), initially introduced by~\citet{treiber2000congested}. Widely accepted in the literature, IDM is a time-continuous car-following model focusing on the interaction between vehicles in a traffic flow through mathematical formulations of vehicle acceleration. Notably, IDM has found practical applications in traffic management systems, autonomous vehicle development, and urban planning, and several modifications and improvements have been proposed over the years \cite{li2015st, debrel2013mod, kesting2010enhanced}.


IDM describes the dynamics of a vehicle $\alpha$ by updating its location $x_\alpha$ and velocity $v_\alpha$ at time $t$ according to the following two ordinary differential equations:
\begin{equation}\label{idm:s}
\dot{x}_\alpha = \frac{dx_\alpha}{dt} = v_\alpha
\end{equation}
\begin{equation}\label{idm:a}
\dot{v}_\alpha= \frac{dv_\alpha}{dt} = a\left[1-\left(\frac{v_\alpha}{v_{0}}\right)^\delta-\left(\frac{s^{\ast}(v_\alpha,\Delta v_\alpha)}{s_\alpha}\right)^2\right], 
\end{equation}
where
\begin{align*}
s^{\ast}(v_\alpha,\Delta v_\alpha)=s_0+v_\alpha T+\frac{v_\alpha \Delta v_\alpha}{2 \sqrt{a b}}.
\end{align*}
$\Delta v_\alpha$ stands for the approaching rate. $a$ is the desired acceleration, $v_0$ the desired speed, $s^{\ast}$ the spacing between the ego and leading vehicle, $s_0$ the minimum spacing, $b$ the comfortable braking deceleration, $T$ desired time headway, and $\delta$ a constant. 

\subsubsection{NGSIM Dataset}

To study human driving behaviors, we leverage the rich and publicly available real-world dataset from the Next Generation Simulation (NGSIM) project~\cite{ngsim}. The dataset contains data collected from four locations in the US: US 101, I-80, Lankershim Boulevard, and Peachtree Street. Unlike other real-world driving datasets, such as those from CARLA, Waymo, and Lyft, which often involve data collection by a limited number of drivers, the NGSIM dataset provides extensive data for thousands of drivers, achieving a much more extensive coverage of human behavior. The vehicle trajectories in the NGSIM dataset have been recorded with a 100 ms time resolution. Each record within the dataset includes the vehicle position in both a local and global coordinate system, lane number, velocity, acceleration, vehicle type, width, and length, and the vehicle ID of the preceding and following vehicle.

\subsection{Complexity of Human Driving Behavior}

We hypothesize that real human driving behavior is inherently unpredictable and stochastic. The limitations of fitting a parametric model, such as the IDM, to human data are evident in its inability to capture the complexity of the behavior. Even more performant predictive models, like the tree-boosting algorithm XGBoost \cite{chen2016xgboost}, face significant challenges in this task. To evaluate this hypothesis, we applied both the IDM and XGBoost to the NGSIM dataset. Initially, we individually fitted an IDM to each driver in the dataset. Subsequently, we fitted a shared IDM to each of the four scenarios in the dataset.

The fitting process of the IDM to the NGSIM dataset involves several key steps. First, we organize the dataset into trajectories, with each trajectory corresponding to an individual driver. Notably, if a driver changes lanes, the trajectory is split into two distinct trajectories. For each trajectory, we extract the initial velocity and the gap to the vehicle in front. Additionally, we include the velocity of the preceding driver as an input to the model. Subsequently, we simulate the acceleration using the IDM equations (see Equations~\ref{idm:a}-\ref{idm:s}) and update the gap to the leading vehicle based on the new velocity and the velocity of the other driver. The simulation occurs with a time step of $dt = 0.1$, corresponding to the duration between frames in NGSIM. The process is akin to solving the IDM ordinary differential equation using the Euler method, and this is repeated for each frame within the trajectory. The optimization objective is to minimize the mean squared error between the predicted velocities and ground truth velocities. The IDM parameters are optimized using Optuna \cite{akiba2019optuna}, with 500 trials per optimization. The parameters subjected to optimization include the desired velocity $v_0$, minimum spacing $s_0$, desired time headway $T$, acceleration $a$, and the comfortable braking deceleration $b$. The exponent $\gamma$ is held constant at four.

The final mean squared errors from our experiment using the NGSIM dataset are presented in Table~\ref{tab:fit_mse}. As expected, fitting an IDM for each driver outperforms the use of shared IDM parameters across the trajectories, with the observed difference being statistically significant. However, it is crucial to highlight the remarkably high standard deviation of the mean squared errors, underscoring the suboptimal nature of the IDM fit.

To qualitatively illustrate the suboptimality of the IDM fit, Figure~\ref{fig:ngsim_idm_velocity} showcases randomly selected trajectories from each NGSIM dataset. The initial states were sampled from the initial states of the trajectories in the NGSIM dataset, and the trajectory was simulated using the learned IDM parameters. Both the trajectories simulated with driver-specific parameters (in green) and those with shared IDM parameters (in orange) fail to match the ground truth data in blue.

Moreover, the intricacy of predicting human driving behavior is further emphasized by the limitations of the tree-boosting algorithm XGBoost in this task \cite{chen2016xgboost}. XGBoost was fitted to the NGSIM data by learning to predict the change in velocity based on the current velocity, the gap to the preceding vehicle in front in seconds and meters, and the velocity difference. No driver-specific variables were included in the training data. Despite training the model for 20,000 tree-boosting iterations, when the training loss had stabilized, Figure~\ref{fig:xgb_onestep_preds} illustrates that the algorithm fails to achieve a low one-step prediction error, even in the supervised setting. Consequently, deploying the learned XGBoost driving policy in the NGSIM environment proves ineffective in predicting realistic driving behaviors, and this issue is exacerbated by accumulating errors, as depicted by the red curves in Figure~\ref{fig:ngsim_idm_velocity}.

It is essential to note that the selected methods, IDM and XGBoost, even struggle to predict the training data, emphasizing the complexity of replicating human behavior. However, it must be acknowledged that the difficulty may be influenced by the lower quality of the NGSIM data that is extracted from raw video files and may contain anomalies. In summary, we contend that sophisticated methods capable of accommodating the stochasticity and complexity of human behavior are indispensable for achieving realistic and accurate human simulations.

\subsection{Diversity of Human Driving Behavior}

We posit that human drivers exhibit unique driving styles, characterized by variations in, for instance, aggression, composure, cornering preferences, and turning behavior. The spectrum of driving styles is extensive. Achieving accuracy and realism in simulations of human driving behavior requires considering and modeling this diversity. Furthermore, we assert that simply sampling, for instance, IDM parameters from a fitted Gaussian may prove inadequate for capturing the nuanced richness and complexity of human driving styles.

To rigorously assess our hypothesis regarding the diversity of human driving styles, we conducted a careful examination of the NGSIM dataset, which provides valuable insights into the behavior of human drivers. Since the NGSIM dataset primarily captures car-following behavior, our initial analysis focuses on three key variables: the average acceleration, the deceleration during braking, and the preferred headway or safety margin to the vehicle in front. 

In Figure~\ref{fig:ngsim_diversity}, we present the distribution of these variables across the four NGSIM datasets using histograms. The first row of subplots depicts the distribution of average acceleration. To derive this, we computed the change in velocity between two consecutive frames for each driver, including pairs of frames where the velocity increased by at least two meters per second. Then, we computed the mean acceleration and plotted the result. Notably, the Peachtree dataset exhibited a lower percentage (approximately 50 \%) of drivers with such accelerations, in contrast to the US 101, I-80, and Lankershim datasets, where at least 98 \% of the drivers demonstrated these types of accelerations. This discrepancy should be considered in the interpretation of our plots. The Peachtree and Lankershim datasets show evidence that the distribution of mean accelerations resembles a multimodal one, whereas the histograms for US 101 and I-80 are unimodal.

The second row of Figure~\ref{fig:ngsim_diversity} displays the distribution of mean decelerations, following a similar process. Pairs of frames where the speed decreased by at least two meters per second were filtered, and the average deceleration for each driver was computed and plotted.

The final row presents plots of the minimum time headway for each driver. This was determined by dividing the distance to the vehicle in front by the current velocity of the ego vehicle, with the minimum value taken for each trajectory. Drivers with a minimum headway of over five seconds were excluded from our analysis due to insufficient traffic. Notably, anomalies in the dataset may contribute to apparent peaks at zero seconds, but we believe these are not representative of actual human behavior.

The modeling of human driving behavior often employs parametric systems such as IDM. In the second part of our analysis of driving style diversity, we fitted an IDM for each individual driver and compared the distribution of IDM parameters across the set of human drivers. Figure ~\ref{fig:ngsim_idm_diversity} illustrates the distribution of the five IDM parameters for each of the four NGSIM datasets.

In the visual representation, the black vertical line signifies the parameters' values when a single model is fitted to every trajectory in the dataset, essentially representing the human mean. However, it's also crucial to account for the inherent noise in fitting the IDM parameters. To estimate this noise, we conducted multiple fittings of IDM for each diver and computed the average standard deviation of the parameter values. The red-shaded area in the figure represents the mean plus-minus two standard deviations. If the distribution of IDM parameters adhered to a Gaussian distribution, and each human driver could be adequately represented with the same set of parameter values, we would expect 95 \% of the values to fall within the shaded region. However, this is not the case, indicating significant diversity in the driving styles.

It is important to note that, as demonstrated in the previous section, the fit of IDM to the data is suboptimal, particularly for the maximum acceleration parameter (a), as illustrated also by Figure~\ref{fig:ngsim_idm_diversity}. This limitation must be considered when drawing conclusions from the experiment.

In summary, our findings provide compelling evidence of the diverse nature of human driving styles. When modeling these tyles using parametric models, our results reveal that the distribution of parameters can exhibit multimodality or fat-tailed characteristics. This implies that simplistic parameter sampling from a Gaussian distribution is unrealistic and should be eschewed. Instead, in realistic simulations, it is advisable to employ more sophisticated sampling strategies to better capture the nuances of human driving behavior.

\subsection{Consistency of Human Driving Behavior}

We hypothesize that human driving behavior is consistent, and this consistency should be accounted for in realistic simulations of human driving. Each human driver has a specific style and will mostly adhere to that, although within the constraints of the natural variability of human behavior. This consistency will also be seen during trajectories; a human will drive similarly at the beginning of a trajectory as at the end. To evaluate this hypothesis, we conducted an experiment with the NGSIM dataset and IDM. We select trajectories from all four datasets in the NGSIM data and split them into half. Then, we fit an IDM to the first and second halves of the trajectory as described in the earlier sections of this paper. Then, we compare if the optimization procedure found the same set of parameters for the two halves. We do this by calculating the Euclidean distance (L2-norm) between the IDM parameters.

Figure~\ref{fig:ngsim_consistency} shows the results for the human consistency experiment. The red curve in the subfigures illustrates the inherent randomness in fitting the IDM parameters. If the parameters are optimized for each trajectory twice, the mean Euclidean distance between the found IDM parameters is approximately 15. The green curve shows the mean Euclidean distance between IDM parameters for two different drivers, which is, as expected, significantly higher than the distance between the parameters for the same drivers. The blue curves in the graphs show that the Euclidean distance between the IDM parameters for the two halves of the same trajectory is lower than the distance between the IDM parameters for different drivers, especially in the case of longer trajectories, which implies that the IDM parameters capture the individual driving style of each driver and, moreover, that this style is consistent over time. If we focus on the longest trajectories in the dataset, this difference in distances is statistically significant. However, this consistency is not visible for shorter trajectories. We believe that it is due to the trajectories being too short and the noise of fitting the IDM parameters starting to dominate the results. Our results support our hypothesis of human driving behavior being temporally consistent, and therefore, it should be incorporated in realistic simulations of human driving behavior.

\section{Conclusion}
In this work, we aim to address the importance of human behavior modeling with specific applications to build high-fidelity data-driven simulators for autonomous driving. We posited the development of a data-driven simulation environment as a learning problem with open-ended solutions focused on theory-based and data-driven models. Our investigation into the complexity, diversity, variability, and temporal consistency of human driving behavior is grounded in empirical analysis using the real-world NGSIM dataset. 
\par
Our key findings reveal that human behaviors are complex, highlighting that theory-based models, like the Intelligent Driver Model (IDM), and data-driven models that fail to explicitly incorporate human driving styles, such as XGBoost, encounter limitations in fully capturing this complexity. Similarly, simplistic parameter sampling may not capture the subtle diversity of human behavior. In addition, our experiments showcased that human driving behavior exhibits temporal consistency, meaning individual driving styles remain stable over time. This consistency is evident from the analysis of IDM parameters across different trajectory halves, emphasizing the need to incorporate this aspect into simulations. 
\par
This research contributes to understanding human driving behavior in the context of autonomous vehicle simulation and paves the way for future explorations. It opens up avenues for formalizing data-driven models to emulate human behavior more effectively and sets the stage for developing models that capture human driving patterns' inherent variability and diversity. These advancements are crucial for creating better human models for simulation.

\bibliography{main}

\end{document}